# Fast RodFIter for Attitude Reconstruction from Inertial Measurements

Yuanxin Wu, *Senior Member, IEEE,* Qi Cai, and Tnieu-Kien Truong, *Life Fellow, IEEE*

***Abstract*—Attitude computation is of vital importance for a variety of applications. Based on the functional iteration of the Rodrigues vector integration equation, the RodFIter method can be advantageously applied to analytically reconstruct the attitude from discrete gyroscope measurements over the time interval of interest. It is promising to produce ultra-accurate attitude reconstruction. However, the RodFIter method imposes high computational load and does not lend itself to onboard implementation. In this paper, a fast approach to significantly reduce RodFIter's computation complexity is presented while maintaining almost the same accuracy of attitude reconstruction. It reformulates the Rodrigues vector iterative integration in terms of the Chebyshev polynomial iteration. Due to the excellent property of Chebyshev polynomials, the fast RodFIter is achieved by means of appropriate truncation of Chebyshev polynomials, with provably guaranteed convergence. Moreover, simulation results validate the speed and accuracy of the proposed method.**

## I. INTRODUCTION

Attitude information is required in many areas, including but not limited to unmanned vehicle navigation and control, virtual/augmented reality, satellite communication, robotics, and computer vision [2]. Attitude computation by integrating angular velocity, e.g., measured by gyroscopes, is an important way to acquire the attitude [3, 4]. For applications with high quality gyroscopes or highly dynamic angular motions, it is important to employ an accurate attitude integration method that can mitigate attitude errors as much as possible, for instance in the context of long-duration GPS-denied inertial navigation. The modern-day attitude algorithm structure in the inertial navigation field, established in 1970s [5, 6], has relied on a simplified rotation vector differential equation for incremental attitude update. It might be argued that the common numerical methods such as the Runge Kutta can also be used to implement the attitude integration although they have been compared infavorably with the modern-day attitude algorithm in early works. Inherently, the above-mentioned methods cannot drive the non-commutativity error to zero with pratically finite sampling intervals. It has long been believed that the modern-day attitude algorithm is already good enough for practical navigation applications [7, 8]. However, in addition to the actual military projectiles with complex high-speed rotations, the cold-atom interference gyroscopes of ultra-high precision are on the horizon as well. The demand for more accurate attitude algorithms becomes increasingly imminent.

Recently, independent works on high-accurate attitude algorithms have appeared or been under way [1, 9]. They share the same spirit of accurately solving the attitude kinematic equation using the fitted angular velocity polynomial function. The main difference among these works is the chosen attitude parameterization. The quaternion is employed in [9] and the Picard-type integration method is used to deal with the quaternion differential equation. Alternatively, the direction cosine matrix (DCM) could be used instead. These methods can be traced back to the Russian work in 1990s [10], where they were respectively classified as Type I and Type II methods. Specifically, the Type II method used the rotation vector instead of DCM. The three-component parameterization is minimal and does not need to satisfy the inherent constraints of the redundant-component parameterizations, such as the quaternion (the unit-norm constraint) and the DCM (the orthogonal and +1-determinant constraint). In view of the finite-polynomial-like differential equation of the three-component Rodrigues vector, Wu [1] proposes the RodFIter method to reconstruct the attitude, which makes a natural use of the iterative function integration of the Rodrigues vector's kinematic equation. It highlights the capability of analytical attitude reconstruction and provably converges to the true attitude if only the angular velocity is exact. The RodFIter method also gives birth to an ultimate attitude algorithm scheme and can be naturally extended to the general rigid motion computation. Specifically, for the sake of better numerical stability, the angular velocity function is fitted by the Chebyshev polynomial instead of the normal polynomial. The RodFIter method is lately found to be related to the Chebyshev-Picard iteration method for the orbital motion solution of initial value and boundary value problems [11-13]. Besides originating from very different background, one of their major differences is that the former has to reconstruct the angular velocity from uniformly-sampled sensor measurements, but the latter is free to use specially-designated cosine sampling for the analytical integrand.

Unfortunately, all high-accurate attitude algorithms face the problem of huge computational burden, especially for real-time applications. Hence, to reduce the complexity, this paper fully exploits the excellent property of Chebyshev polynomial

This work was supported in part by National Natural Science Foundation of China (61422311, 61673263), Joint Fund of China Ministry of Education (6141A02022309) and Hunan Provincial Natural Science Foundation of China (2015JJ1021). A short summary was presented the symposium on Inertial Sensors and Systems (ISS), Braunschweig, Germany, 2018.

Authors' address: Shanghai Key Laboratory of Navigation and Location-based Services, School of Electronic Information and Electrical Engineering, Shanghai Jiao Tong University, Shanghai, China, 200240, E-mail: (yuanx_wu@hotmail.com).

to significantly improve the computational efficiency of RodFIter. The idea also applies to other high-accurate attitude algorithms. In addition, it is proved that the fast RodFIter also converges to the true attitude in terms of finite polynomials. The remainder of the paper is organized as follows: Section II briefly reviews the RodFIter method. Section III is devoted to the computational complexity analysis of RodFIter and reformulates the iterative integration of the Rodrigues vector as the iterative computation of Chebyshev polynomial coefficients. Section IV proposes the fast version of RodFIter by appropriate polynomial truncation, and analyzes its convergence property and error characteristics. Section V evaluates the fast RodFIter contrasting the original version by numerical simulations. A brief summary is given in the last section of the paper.

## II. Brief review of RodFIter

The following is a brief review of the RodFIter method. For more detail, the interested reader is referred to [1].

The RodFIter method computes the incremental Rodrigues vector over the time interval $[0 \quad t]$ by

$$\mathbf{g}_{l+1} = \int_0^t \left( \mathbf{I}_3 + \frac{1}{2}\mathbf{g}_l \times + \frac{1}{4}\mathbf{g}_l \mathbf{g}_l^T \right) \boldsymbol{\omega}\, dt, \quad l \ge 0, \tag{1}$$

with the initial Rodrigues vector given by $\mathbf{g}_0 = 0$. Here, the angular velocity function $\boldsymbol{\omega}$ may be fitted by a polynomial function from discrete gyroscope measurements.

The RodFIter method has a provable convergence as stated below.

*Theorem 1*: Given the true angular velocity function $\boldsymbol{\omega}$ over the interval $[0 \quad t]$, the iterative process as given in (1) converges to the true Rodrigues vector function when $t \sup|\boldsymbol{\omega}| < 2$.

The operator $|\cdot|$ denotes the magnitude of a vector. For proof details, see [1].

In view of the fact that the Chebyshev polynomial is a sequence of orthogonal polynomial bases and has better numerical stability than the normal polynomial [14], Wu [1] uses the former to fit the discrete angular velocity or angular increment measurements. The Chebyshev polynomial of the first kind is defined over the interval $[-1 \quad 1]$ by the recurrence relation as

$$F_0(x) = 1,\ F_1(x) = x,\ F_{i+1}(x) = 2xF_i(x) - F_{i-1}(x), \tag{2}$$

where $F_i(x)$ is the $i^{\text{th}}$-degree Chebyshev polynomial of the first kind.

Now, let the discrete measurement time instants and the considered time interval be denoted by $t_k$ and $[0 \quad t_N]$, respectively. For each attitude update using $N$ samples, a number of angular velocity measurements $\tilde{\boldsymbol{\omega}}_{t_k}$ or angular increment measurements $\Delta\tilde{\boldsymbol{\theta}}_{t_k}$ for ($k = 1, 2, \ldots N$) are used to fit the angular velocity function. In order to apply the Chebyshev polynomials, the actual time interval is mapped onto $[-1 \quad 1]$ by letting $t = \frac{t_N}{2}(1+\tau)$. Then the angular velocity over the mapped interval is fitted by the Chebyshev polynomial in time up to the degree of $N$-1, given by

$$\hat{\boldsymbol{\omega}} = \sum_{i=0}^{n} \mathbf{c}_i F_i(\tau),\ n \le N-1. \tag{3}$$

The coefficient $\mathbf{c}_i$ in (3) is determined for the case of angular velocity measurement by solving the equation as follows:

$$\mathbf{A}_\omega \begin{bmatrix} \mathbf{c}_0^T \\ \mathbf{c}_1^T \\ \vdots \\ \mathbf{c}_n^T \end{bmatrix} \triangleq \begin{bmatrix} 1 & F_1(\tau_1) & \cdots & F_n(\tau_1) \\ 1 & F_1(\tau_2) & \cdots & F_n(\tau_2) \\ \vdots & \vdots & \vdots & \vdots \\ 1 & F_1(\tau_N) & \cdots & F_n(\tau_N) \end{bmatrix} \begin{bmatrix} \mathbf{c}_0^T \\ \mathbf{c}_1^T \\ \vdots \\ \mathbf{c}_n^T \end{bmatrix} = \begin{bmatrix} \tilde{\boldsymbol{\omega}}_{t_1}^T \\ \tilde{\boldsymbol{\omega}}_{t_2}^T \\ \vdots \\ \tilde{\boldsymbol{\omega}}_{t_N}^T \end{bmatrix}. \tag{4}$$

For $N$ angular increment measurements $\Delta\tilde{\boldsymbol{\theta}}_{t_k}$ for $(k = 1, 2, \ldots N)$, the angular velocity is also fitted by the Chebyshev polynomial in time up to the order of $N$-1. According to the integral property of the Chebyshev polynomial [14], we have

$$G_{i,[\tau_{k-1}\ \tau_k]} \triangleq \int_{\tau_{k-1}}^{\tau_k} F_i(\tau)\, d\tau = \begin{cases} \left( \dfrac{iF_{i+1}(\tau_k)}{i^2-1} - \dfrac{\tau_k F_i(\tau_k)}{i-1} \right) - \left( \dfrac{iF_{i+1}(\tau_{k-1})}{i^2-1} - \dfrac{\tau_{k-1} F_i(\tau_{k-1})}{i-1} \right), & i \ne 1 \\ \dfrac{\tau_k^2 - \tau_{k-1}^2}{2}, & i = 1. \end{cases} \tag{5}$$

With the aid of (3) and (5), the angular increment related to the fitted angular velocity is given by

$$\begin{aligned} \Delta\hat{\boldsymbol{\theta}}_{t_k} &= \int_{t_{k-1}}^{t_k} \hat{\boldsymbol{\omega}}\, dt \overset{t = t_N(1+\tau)/2}{=} \frac{t_N}{2} \int_{\tau_{k-1}}^{\tau_k} \hat{\boldsymbol{\omega}}\, d\tau \\ &= \frac{t_N}{2} \sum_{i=0}^{n} \mathbf{c}_i \int_{\tau_{k-1}}^{\tau_k} F_i(\tau)\, d\tau = \frac{t_N}{2} \sum_{i=0}^{n} \mathbf{c}_i G_{i,[\tau_{k-1}\ \tau_k]}. \end{aligned} \tag{6}$$

The coefficient $\mathbf{c}_i$ in (6) is determined for the case of angular increment measurement by solving the following equation:

$$\begin{aligned} \mathbf{A}_\theta \begin{bmatrix} \mathbf{c}_0^T \\ \mathbf{c}_1^T \\ \vdots \\ \mathbf{c}_n^T \end{bmatrix} &\triangleq \begin{bmatrix} G_{0,[\tau_0\ \tau_1]} & G_{1,[\tau_0\ \tau_1]} & \cdots & G_{n,[\tau_0\ \tau_1]} \\ G_{0,[\tau_1\ \tau_2]} & G_{1,[\tau_1\ \tau_2]} & \cdots & G_{n,[\tau_1\ \tau_2]} \\ \vdots & \vdots & \vdots & \vdots \\ G_{0,[\tau_{N-1}\ \tau_N]} & G_{1,[\tau_{N-1}\ \tau_N]} & \cdots & G_{n,[\tau_{N-1}\ \tau_N]} \end{bmatrix} \begin{bmatrix} \mathbf{c}_0^T \\ \mathbf{c}_1^T \\ \vdots \\ \mathbf{c}_n^T \end{bmatrix} \\ &= \frac{2}{t_N} \begin{bmatrix} \Delta\tilde{\boldsymbol{\theta}}_{t_1}^T \\ \Delta\tilde{\boldsymbol{\theta}}_{t_2}^T \\ \vdots \\ \Delta\tilde{\boldsymbol{\theta}}_{t_N}^T \end{bmatrix}. \end{aligned} \tag{7}$$

With the converging condition of Theorem 1 in mind, substituting (3) into the iteration process (1) is supposed to well reconstruct the Rodrigues vector function as a finite polynomial, as the group of polynomials are closed under elementary arithmetic operations. It is an appreciated benefit due to the simplicity of the Rodrigues vector's differential function. Analytic development of the iterative process in (1) is tedious so that Wu's work [1] turns to the symbolic computation toolbox of Matlab to implement the iterative function integration (1). However, the complexity of symbolic computation is mountainous and on the Matlab platform the RodFIter method is several thousand times bigger than the mainstream attitude algorithm in computation load [1]. It results in a nontrivial problem for real-time applications and thus should be overcome by a more efficient algorithm for software or hardware implementation.

## III. RodFIter in Terms of Chebyshev Polynomial Iterative Computation

Next we will reformulate the RodFIter method as the iterative computation of Chebyshev polynomial coefficients and perform the analysis of computational complexity.

If the angular velocity is smooth, the absolute values of Chebyshev coefficients $\mathbf{c}_i$ in (3) will decrease exponentially [14]. Assume the Rodrigues vector at the $l$-th iteration is given by a weighted sum of Chebyshev polynomials, say

$$\mathbf{g}_l \triangleq \sum_{i=0}^{m_l} \mathbf{b}_{l,i} F_i(\tau), \tag{8}$$

where $m_l$ is the maximum degree and $\mathbf{b}_{l,i}$ is the coefficient of $i^{\text{th}}$-degree Chebyshev polynomial at the $l$-th iteration. The integral over $[0 \quad t]$ in (1) is transformed to that over the mapped interval of Chebyshev polynomials, that is,

$$\begin{aligned}\mathbf{g}_{l+1} &= \int_0^t \left( \mathbf{I}_3 + \frac{1}{2}\mathbf{g}_l \times + \frac{1}{4}\mathbf{g}_l\mathbf{g}_l^T \right) \boldsymbol{\omega}\, dt \\ &= \frac{t_N}{2}\left( \int_{-1}^{\tau} \boldsymbol{\omega}\, d\tau + \frac{1}{2}\int_{-1}^{\tau} \mathbf{g}_l \times \boldsymbol{\omega}\, d\tau + \frac{1}{4}\int_{-1}^{\tau} \mathbf{g}_l\mathbf{g}_l^T \boldsymbol{\omega}\, d\tau \right).\end{aligned} \tag{9}$$

Substituting (3), the first subintegral on the right side of (9) has the form

$$\int_{-1}^{\tau} \boldsymbol{\omega}\, d\tau = \sum_{i=0}^{n} \mathbf{c}_i \int_{-1}^{\tau} F_i(\tau)\, d\tau = \sum_{i=0}^{n} \mathbf{c}_i G_{i,[-1\ \tau]}. \tag{10}$$

For any $j,k \ge 0$, the Chebyshev polynomial of first kind satisfies the equality [14] as follows:

$$F_j(\tau)F_k(\tau) = \frac{1}{2}\left( F_{j+k}(\tau) + F_{|j-k|}(\tau) \right). \tag{11}$$

Then according to (2), (5) and (11), the integrated $i^{\text{th}}$-degree Chebyshev polynomial can be expressed as a combination of Chebyshev polynomials, given by

$$G_{i,[-1\ \tau]} = \int_{-1}^{\tau} F_i(\tau)\, d\tau$$
$$= \begin{cases} \left( \dfrac{iF_{i+1}(\tau)}{i^2-1} - \dfrac{\tau F_i(\tau)}{i-1} \right) - \left( \dfrac{iF_{i+1}(-1)}{i^2-1} + \dfrac{F_i(-1)}{i-1} \right) F_0(\tau) \\ = \left( \dfrac{iF_{i+1}(\tau)}{i^2-1} - \dfrac{F_{i+1}(\tau) + F_{|i-1|}(\tau)}{2(i-1)} \right) - \left( \dfrac{iF_{i+1}(-1)}{i^2-1} + \dfrac{F_i(-1)}{i-1} \right) F_0(\tau) \\ = \left( \dfrac{F_{i+1}(\tau)}{2(i+1)} - \dfrac{F_{|i-1|}(\tau)}{2(i-1)} \right) - \dfrac{(-1)^i}{i^2-1} F_0(\tau) \quad \text{for } i \ne 1, \\ \dfrac{\tau^2-1}{2} = \dfrac{F_{i+1}(\tau)}{4} - \dfrac{F_0(\tau)}{4} \quad \text{for } i = 1, \end{cases} \tag{12}$$

where $F_i(-1) = (-1)^i$.

Evidently, the middle integral term on the right side of (9) is

$$\begin{aligned}\int_{-1}^{\tau} \mathbf{g}_l \times \boldsymbol{\omega}\, d\tau &= \int_{-1}^{\tau} \left[ \sum_{i=0}^{m_l} \mathbf{b}_{l,i} F_i(\tau) \right] \times \sum_{j=0}^{n} \mathbf{c}_j F_j(\tau) d\tau \\ &= \sum_{i=0}^{m_l}\sum_{j=0}^{n} \mathbf{b}_{l,i} \times \mathbf{c}_j \int_{-1}^{\tau} F_i(\tau)F_j(\tau)\, d\tau \\ &= \frac{1}{2}\sum_{i=0}^{m_l}\sum_{j=0}^{n} \mathbf{b}_{l,i} \times \mathbf{c}_j \int_{-1}^{\tau} \left( F_{i+j}(\tau) + F_{|i-j|}(\tau) \right) d\tau \\ &= \frac{1}{2}\sum_{i=0}^{m_l}\sum_{j=0}^{n} \mathbf{b}_{l,i} \times \mathbf{c}_j \left( G_{i+j,[-1\ \tau]} + G_{|i-j|,[-1\ \tau]} \right).\end{aligned} \tag{13}$$

And the last integral term on the right side of (9) is found to be

$$\begin{aligned}\int_{-1}^{\tau} \mathbf{g}_l\mathbf{g}_l^T \boldsymbol{\omega}\, d\tau &= \int_{-1}^{\tau} \left[ \sum_{i=0}^{m_l} \mathbf{b}_{l,i} F_i(\tau) \right] \left[ \sum_{j=0}^{m_l} \mathbf{b}_{l,j}^T F_j(\tau) \right] \left[ \sum_{k=0}^{n} \mathbf{c}_k F_k(\tau) \right] d\tau \\ &= \sum_{i=0}^{m_l}\sum_{j=0}^{m_l}\sum_{k=0}^{n} \mathbf{b}_{l,i}\mathbf{b}_{l,j}^T \mathbf{c}_k \int_{-1}^{\tau} F_i(\tau)F_j(\tau)F_k(\tau)\, d\tau \\ &= \frac{1}{2}\sum_{i=0}^{m_l}\sum_{j=0}^{m_l}\sum_{k=0}^{n} \mathbf{b}_{l,i}\mathbf{b}_{l,j}^T \mathbf{c}_k \int_{-1}^{\tau} \left( F_{i+j}(\tau) + F_{|i-j|}(\tau) \right) F_k(\tau)\, d\tau \\ &= \frac{1}{4}\sum_{i=0}^{m_l}\sum_{j=0}^{m_l}\sum_{k=0}^{n} \mathbf{b}_{l,i}\mathbf{b}_{l,j}^T \mathbf{c}_k \int_{-1}^{\tau} \begin{pmatrix} F_{i+j+k}(\tau) + F_{|i+j-k|}(\tau) \\ + F_{|i-j|+k}(\tau) + F_{||i-j|-k|}(\tau) \end{pmatrix} d\tau \\ &= \frac{1}{4}\sum_{i=0}^{m_l}\sum_{j=0}^{m_l}\sum_{k=0}^{n} \mathbf{b}_{l,i}\mathbf{b}_{l,j}^T \mathbf{c}_k \begin{pmatrix} G_{i+j+k,[-1\ \tau]} + G_{|i+j-k|,[-1\ \tau]} \\ + G_{|i-j|+k,[-1\ \tau]} + G_{||i-j|-k|,[-1\ \tau]} \end{pmatrix}.\end{aligned} \tag{14}$$

It is of interest to note that the integrated $i^{\text{th}}$-degree Chebyshev polynomial in (12), namely $G_{i,[-1\ \tau]}$, is a Chebyshev polynomial of degree $i$+1. This means that the integral terms (10), (13) and (14) are all Chebyshev polynomials, so is the Rodrigues vector in (9). A substitution of (10), (13) and (14) into (9) yields

$$\begin{aligned}\mathbf{g}_{l+1} &= \frac{t_N}{2}\begin{pmatrix} \displaystyle\sum_{i=0}^{n} \mathbf{c}_i G_{i,[-1\ \tau]} + \frac{1}{4}\sum_{i=0}^{m_l}\sum_{j=0}^{n} \mathbf{b}_{l,i} \times \mathbf{c}_j \left( G_{i+j,[-1\ \tau]} + G_{|i-j|,[-1\ \tau]} \right) \\ \displaystyle + \frac{1}{16}\sum_{i=0}^{m_l}\sum_{j=0}^{m_l}\sum_{k=0}^{n} \mathbf{b}_{l,i}\mathbf{b}_{l,j}^T \mathbf{c}_k \begin{pmatrix} G_{i+j+k,[-1\ \tau]} + G_{|i+j-k|,[-1\ \tau]} \\ + G_{|i-j|+k,[-1\ \tau]} + G_{||i-j|-k|,[-1\ \tau]} \end{pmatrix} \end{pmatrix} \\ &\triangleq \sum_{i=0}^{m_{l+1}} \mathbf{b}_{l+1,i} F_i(\tau),\end{aligned} \tag{15}$$

where $m_{l+1} = 2m_l + n + 1$ with $m_0 = 0$.

It can be readily verified that $m_l = (2^l - 1)(n+1)$. The computational complexity of (15) in terms of the number of weighted terms is proportional to $O(nm_l^2) = O\left( (2^l-1)^2 n(n+1)^2 \right)$. For instance, for $n = 8$ at the $7^{\text{th}}$ iteration, $m_l = 1143$ and the computational complexity is proportional to $O(10^7)$. Although this complexity analysis is based on the Chebyshev polynomial representation, it is comparable to that of the original RodFIter in [1].

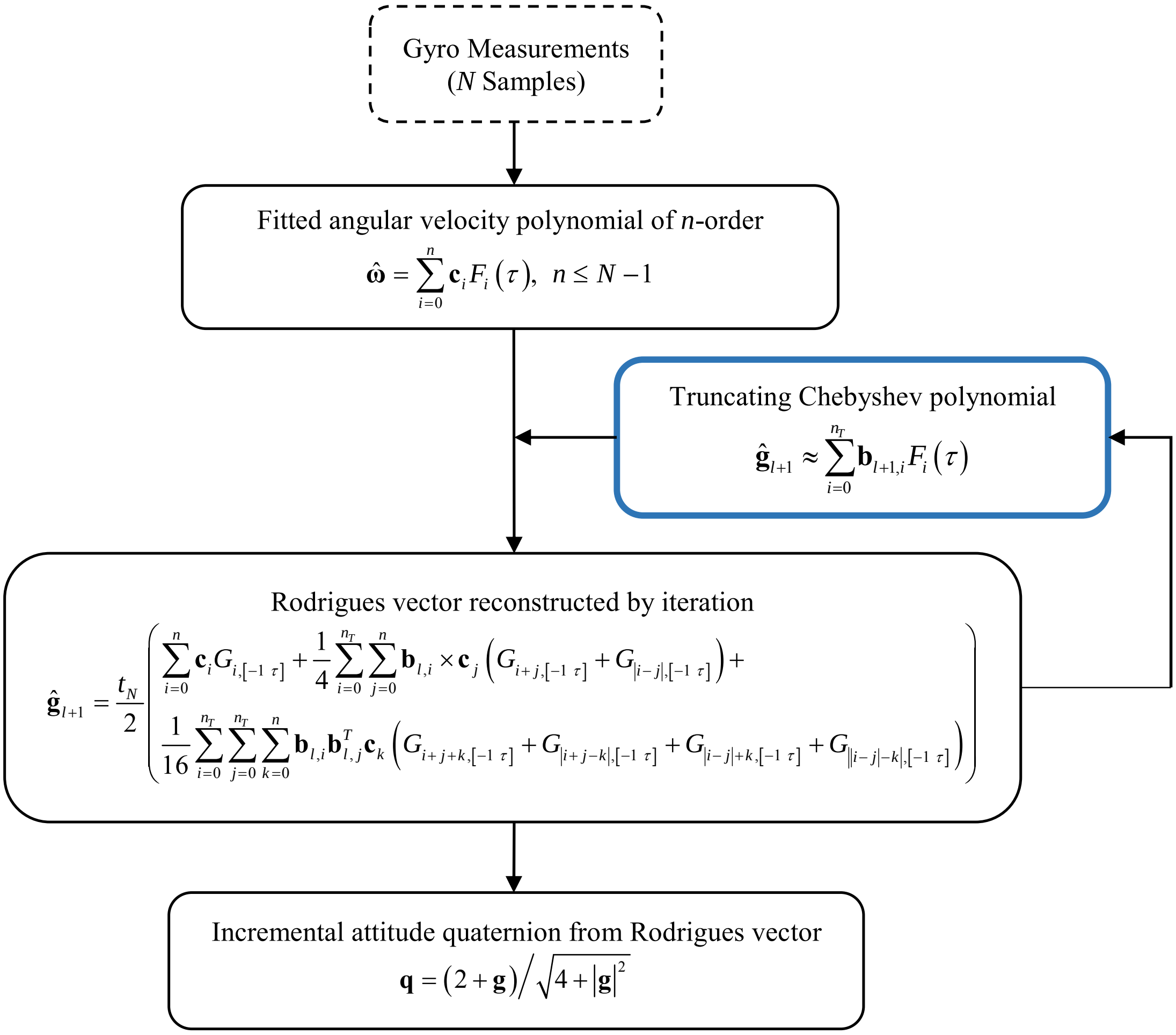

Figure 1. Flowchart of fast RodFIter. Chebyshev polynomial truncation is highlighted by thick-lined blue rectangle.

## IV. Fast RodFIter by Polynomial Truncation

As the iteration goes on, the degree of the Chebyshev polynomials increases quickly and poses huge computational burden. This section will reduce the computational burden by means of polynomial truncation.

### *A. Fast RodFIter and Its Convergence Property*

Recalling (1), the angular velocity is fitted by a Chebyshev polynomial of degree $n$ in time and thus the first subintegral of (9) is accurate up to degree $n+1$. It is reasonable to abandon those polynomials of higher order than some threshold so that much computation could be saved.

At each iteration, suppose the Chebyshev polynomial of the Rodrigues vector is truncated up to degree $n_T\ (\ge n+1)$, i.e.,

$$\hat{\mathbf{g}}_l = \sum_{i=0}^{n_T} \mathbf{b}_{l,i} F_i(\tau). \quad (16)$$

According to the coefficient-decreasing and $\pm 1$-bounded properties of Chebyshev polynomials [14], the truncation error is bounded by the coefficient of the first neglected Chebyshev polynomial, namely, $\left|\delta\mathbf{g}_l^t\right| \le \left|\mathbf{b}_{l,n_T+1}\right|$. The superscript '$t$' denotes that the error is owed to the polynomial truncation. The iteration (15) becomes

$$\hat{\mathbf{g}}_{l+1} = \frac{t_N}{2}\left( \begin{array}{l} \sum_{i=0}^{n} \mathbf{c}_i G_{i,[-1\ \tau]} + \frac{1}{4}\sum_{i=0}^{n_T}\sum_{j=0}^{n} \mathbf{b}_{l,i}\times\mathbf{c}_j \left(G_{i+j,[-1\ \tau]} + G_{|i-j|,[-1\ \tau]}\right) \\ + \frac{1}{16}\sum_{i=0}^{n_T}\sum_{j=0}^{n_T}\sum_{k=0}^{n} \mathbf{b}_{l,i}\mathbf{b}_{l,j}^T\mathbf{c}_k \left(\begin{array}{l} G_{i+j+k,[-1\ \tau]} + G_{|i+j-k|,[-1\ \tau]} \\ + G_{|i-j|+k,[-1\ \tau]} + G_{||i-j|-k|,[-1\ \tau]}\end{array}\right)\end{array}\right)$$
$$\triangleq \sum_{i=0}^{2n_T+n+1} \mathbf{b}_{l+1,i}F_i(\tau) \overset{\text{polynomial truncation}}{\approx} \sum_{i=0}^{n_T}\mathbf{b}_{l+1,i}F_i(\tau), \quad (17)$$

where the last approximation is due to the truncation at each iteration. Figure 1 presents the flowchart of the fast RodFIter. By way of polynomial truncation, the computational complexity is reduced to $O(nn_T^2)$. For instance, it will be proportional to $O(10^3)$ for $n_T = n+1$ and $n = 8$ at each iteration. Compared with the non-truncated RodFIter (15) at the 7th iteration, the computational burden is remarkably mitigated by over 10,000 times.

*Theorem 2*: Given the true angular velocity function $\boldsymbol{\omega}$ over the interval $[0 \quad t]$, the fast RodFIter (17) converges to the true Rodrigues vector up to the truncated polynomial degree $n_T$ when $t\sup|\boldsymbol{\omega}|<2$.

Proof. Assume $\hat{\mathbf{g}}_l = \sum_{i=0}^{n_T}\mathbf{b}_{l,i}F_i(\tau)$ for the current iteration $l\ (=1,2,\ldots)$ of the fast RodFIter. According to Theorem 1, the original RodFIter, taking $\hat{\mathbf{g}}_l$ as the initial Rodrigues vector, namely $\mathbf{g}_0 \triangleq \hat{\mathbf{g}}_l$, converges to the true Rodrigues vector when $t\sup|\boldsymbol{\omega}|<2$. For the first iteration of the original RodFIter, $\mathbf{g}_1 = \sum_{i=0}^{2n_T+n+1}\mathbf{b}_{l+1,i}F_i(\tau)$, where the coefficients are defined in (17). According to the proof in Theorem 1, $\mathbf{g}_1$ is closer to the true Rodrigues vector than $\mathbf{g}_0$. Thus, for the fast RodFIter, $\hat{\mathbf{g}}_{l+1} = \sum_{i=0}^{n_T}\mathbf{b}_{l+1,i}F_i(\tau)$ is a better approximation of the true Rodrigues vector than $\hat{\mathbf{g}}_l$.

The analysis mentioned above applies to each iteration, so the fast RodFIter (17) converges to the true Rodrigues vector up to the truncated polynomial degree $n_T$.

■

In analogy with Theorem 2 in [1], the above result can be extended to the case of erroneous angular velocity and is summarized in the evident theorem below.

*Theorem 3*: Given the error-contaminated angular velocity function $\hat{\boldsymbol{\omega}} = \boldsymbol{\omega} + \delta\boldsymbol{\omega}$ over the interval $[0 \quad t]$, the fast RodFIter (17) converges to the corresponding Rodrigues vector up to the truncated polynomial degree $n_T$ when $t\sup|\hat{\boldsymbol{\omega}}|<2$.

It should be noted hereafter that the angular velocity error $\delta\boldsymbol{\omega}$ may include the gyroscope error as well as the polynomial fitting error in (8).

## *B. Error Analysis*

It is obvious to see that the fast RodFIter has three error sources: the angular velocity's error $\delta\boldsymbol{\omega}$, the initial error, and the truncation error at the *l*-th iteration $\delta\mathbf{g}_l^t$. Next consider the fact that the Rodrigues vector in each attitude update interval is normally a small quantity. The Rodrigues vector error (to first order) of the fast RodFIter propagates as

$$\begin{aligned}
\delta\hat{\mathbf{g}}_{l+1} &= \delta\int_0^t\left(\mathbf{I}_3 + \frac{1}{2}\hat{\mathbf{g}}_l \times + \frac{1}{4}\hat{\mathbf{g}}_l\hat{\mathbf{g}}_l^T\right)\boldsymbol{\omega}\,dt + \delta\mathbf{g}_{l+1}^t \\
&= \int_0^t \delta\boldsymbol{\omega}\,dt + \frac{1}{2}\int_0^t \delta\hat{\mathbf{g}}_l \times \boldsymbol{\omega}\,dt + \frac{1}{2}\int_0^t \hat{\mathbf{g}}_l \times \delta\boldsymbol{\omega}\,dt \\
&\quad + \frac{1}{4}\int_0^t \delta\hat{\mathbf{g}}_l\hat{\mathbf{g}}_l^T\boldsymbol{\omega}\,dt + \frac{1}{4}\int_0^t \hat{\mathbf{g}}_l\delta\hat{\mathbf{g}}_l^T\boldsymbol{\omega}\,dt + \frac{1}{4}\int_0^t \hat{\mathbf{g}}_l\hat{\mathbf{g}}_l^T\delta\boldsymbol{\omega}\,dt + \delta\mathbf{g}_{l+1}^t \\
&\approx \int_0^t \delta\boldsymbol{\omega}\,dt + \frac{1}{2}\int_0^t \delta\hat{\mathbf{g}}_l \times \boldsymbol{\omega}\,dt + \delta\mathbf{g}_{l+1}^t.
\end{aligned} \tag{18}$$

Thus, we have

$$\begin{aligned}
|\delta\hat{\mathbf{g}}_{l+1}| &\le \left|\int_0^t \delta\boldsymbol{\omega}\,dt\right| + \frac{1}{2}\left|\int_0^t \delta\hat{\mathbf{g}}_l \times \boldsymbol{\omega}\,dt\right| + \left|\delta\mathbf{g}_{l+1}^t\right| \\
&\le t\cdot\sup|\delta\boldsymbol{\omega}| + \sup|\delta\hat{\mathbf{g}}_l|\frac{t\cdot\sup|\boldsymbol{\omega}|}{2} + \left|\delta\mathbf{g}_{l+1}^t\right|,
\end{aligned} \tag{19}$$

where $|\cdot|$ denotes the vector norm. It means

$$\begin{aligned}
&\sup|\delta\hat{\mathbf{g}}_{l+1}| \le t\sup|\delta\boldsymbol{\omega}| + \\
&\quad\left(t\sup|\delta\boldsymbol{\omega}| + \sup|\delta\hat{\mathbf{g}}_{l-1}|\frac{t\sup|\boldsymbol{\omega}|}{2} + \sup\left|\delta\mathbf{g}_l^t\right|\right)\frac{t\sup|\boldsymbol{\omega}|}{2} \\
&\quad + \sup\left|\delta\mathbf{g}_{l+1}^t\right| \\
&= t\sup|\delta\boldsymbol{\omega}|\left(1+\frac{t\sup|\boldsymbol{\omega}|}{2}\right) + \sup|\delta\hat{\mathbf{g}}_{l-1}|\left(\frac{t\sup|\boldsymbol{\omega}|}{2}\right)^2 \\
&\quad + \sup\left|\delta\mathbf{g}_l^t\right|\frac{t\sup|\boldsymbol{\omega}|}{2} + \sup\left|\delta\mathbf{g}_{l+1}^t\right| \\
&\cdots \\
&\le t\sup|\delta\boldsymbol{\omega}|\sum_{i=0}^{l}\left(\frac{t\sup|\boldsymbol{\omega}|}{2}\right)^i + \sup|\delta\hat{\mathbf{g}}_0|\left(\frac{t\sup|\boldsymbol{\omega}|}{2}\right)^{l+1} \\
&\quad + \sum_{i=1}^{l+1}\sup\left|\delta\mathbf{g}_i^t\right|\left(\frac{t\sup|\boldsymbol{\omega}|}{2}\right)^{l+1-i} \\
&\le t\sup|\delta\boldsymbol{\omega}|\frac{1-\left(t\sup|\boldsymbol{\omega}|/2\right)^{l+1}}{1-t\sup|\boldsymbol{\omega}|/2} + \sup|\delta\hat{\mathbf{g}}_0|\left(\frac{t\sup|\boldsymbol{\omega}|}{2}\right)^{l+1} \\
&\quad + \sum_{i=1}^{l+1}\left|\mathbf{b}_{i,n_T+1}\right|\left(\frac{t\sup|\boldsymbol{\omega}|}{2}\right)^{l+1-i}.
\end{aligned} \tag{20}$$

In view of the above equation, the first term and the second term are owed to the angular velocity error and the initial Rodrigues vector error, respectively. Since $t\sup|\boldsymbol{\omega}|<2$ as required by the convergence condition, for large iterations the second term gradually vanishes and the first term approaches

$$t\sup|\delta\boldsymbol{\omega}|\frac{1-\left(t\sup|\boldsymbol{\omega}|/2\right)^{l+1}}{1-t\sup|\boldsymbol{\omega}|/2} \approx \frac{t\sup|\delta\boldsymbol{\omega}|}{1-t\sup|\boldsymbol{\omega}|/2}. \tag{21}$$

It is a slightly looser bound than Proposition 2 in [1]. The third term of (20) is owed to the polynomial truncation at each iteration. Because in practice $t\sup|\boldsymbol{\omega}|/2 \ll 1$, the weights of the early iterations are much smaller than those of later iterations and the third term can be approximated by the last truncation error, i.e.,

$$\sum_{i=1}^{l+1}\left|\mathbf{b}_{i,n_T+1}\right|\left(\frac{t\sup|\boldsymbol{\omega}|}{2}\right)^{l+1-i} \approx \left|\mathbf{b}_{l+1,n_T+1}\right|. \tag{22}$$

Therefore, using (21) and (22), the Rodrigues vector error of the fast RodFIter in (20) can be approximately bounded by

$$\sup|\delta\hat{\mathbf{g}}_{l+1}| \le \frac{t\sup|\delta\boldsymbol{\omega}|}{1-t\sup|\boldsymbol{\omega}|/2} + \left|\mathbf{b}_{l+1,n_T+1}\right| \approx t\sup|\delta\boldsymbol{\omega}| + \left|\mathbf{b}_{l+1,n_T+1}\right|. \tag{23}$$

This indicates that the fast RodFIter's error is generally dominated by the angular velocity error and the last truncation error. An ideal case is when $\delta\boldsymbol{\omega} = 0$, namely the angular velocity fitting error is zero, for which higher order of truncation means higher accuracy.

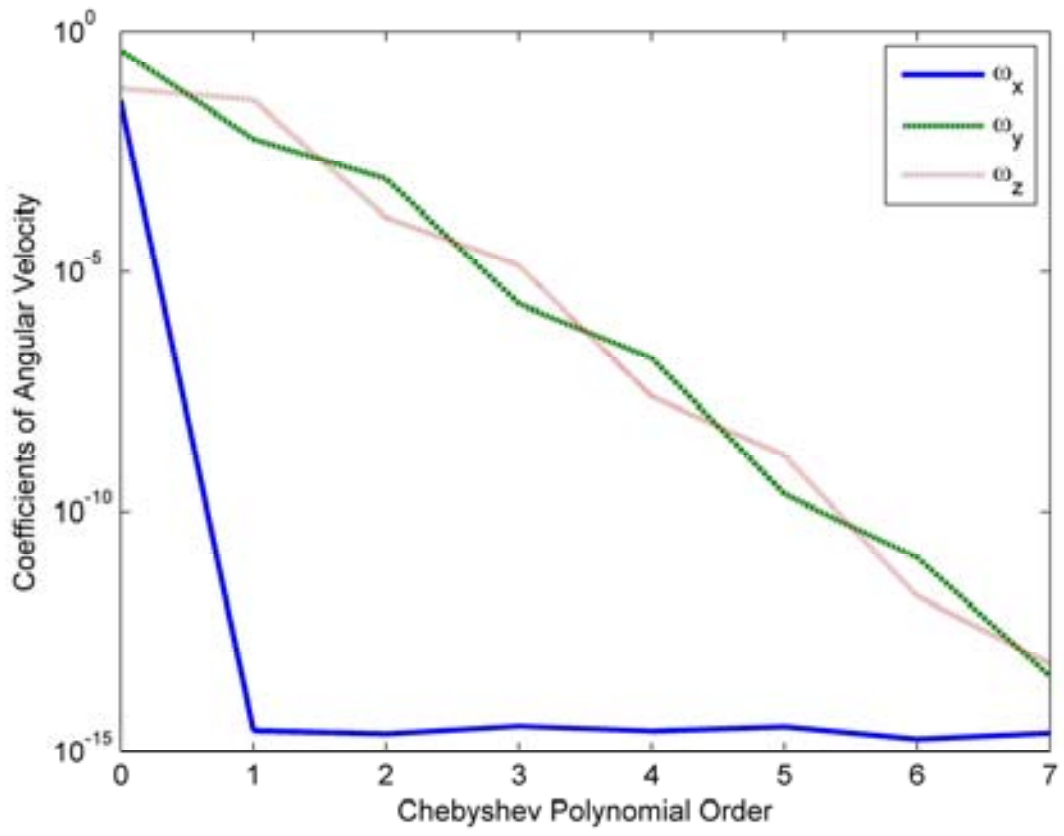

Figure 2. Chebyshev polynomial coefficients (absolute value) of fitted angular velocity (3) for the first update interval (*N*=8).

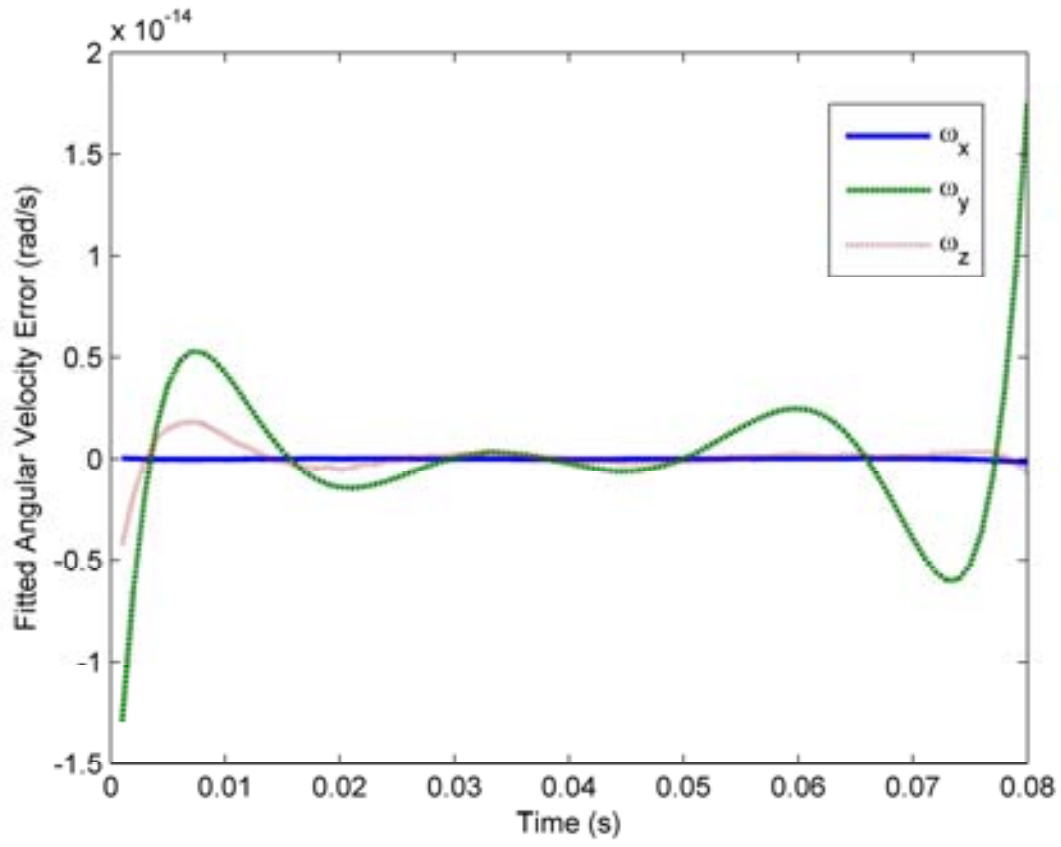

Figure 3. Error of fitted angular velocity Chebyshev polynomial, as compared with the true angular velocity (24), for the first update interval (*N*=8).

For gyroscopes with a priori-known bias $\boldsymbol{\varepsilon}_g$, for example, the polynomial truncation error does not have to be too small, and it is acceptable to just stay below some prescribed percentage, say $\eta$, of that incurred by the gyroscope bias, i.e., $\left|\mathbf{b}_{l,n_T+1}\right| < \eta\left|\boldsymbol{\varepsilon}_g\right| t$. The priori information of the gyroscope error might help decrease the truncation degree $n_T$ and thus further reduce the computational cost.

## V. Simulation Results

Next the coning motion is used to evaluate the fast RodFIter algorithm and compare with the original version [1]. The coning motion has explicit analytical expressions in angular velocity and the associated Rodrigues vector and attitude, so it is widely employed as a standard criterion for algorithm accuracy assessment in the inertial navigation field [3].

The angular velocity of the coning motion is given by

$$\boldsymbol{\omega} = \Omega\left[-2\sin^2(\alpha/2) \quad -\sin(\alpha)\sin(\Omega t) \quad \sin(\alpha)\cos(\Omega t)\right]^T. \tag{24}$$

The corresponding Rodrigues vector is

$$\mathbf{g} = 2\tan\left(\frac{\alpha}{2}\right)\left[0 \quad \cos(\Omega t) \quad \sin(\Omega t)\right]^T, \tag{25}$$

where the coning angle is set to $\alpha = 10\deg$, the coning frequency $\Omega = 0.74\pi$. The attitude quaternion can be obtained by

$$\mathbf{q} = \frac{2+\mathbf{g}}{\sqrt{4+|\mathbf{g}|^2}}. \tag{26}$$

The angular increment measurement from gyroscopes is assumed and the discrete sampling rate is nominally set to 100 Hz. The following angle metric is used to quantify the attitude computation error

$$\varepsilon_{att} = 2\left|\left[\mathbf{q}^* \circ \hat{\mathbf{q}}\right]_{2:4}\right|, \tag{27}$$

where $\hat{\mathbf{q}}$ denotes the quaternion estimate from the Rodrigues vector, and $[\cdot]_{2:4}$ is the sub-vector formed by the last three elements of error quaternion. Hereafter the order of the fitted angular velocity is uniformly set to $n = N-1$.

Figure 2 plots the absolute values of the Chebyshev polynomial coefficients of the fitted angular velocity in (3) for the first update interval when $N = 8$. The magnitude-decreasing property of the Chebyshev polynomial coefficient is apparent. The Runge effect is clearly observed at both ends and might be depressed to some extent by means of using data samples in the neighboring time intervals. It is supposed to further improve the accuracy of the RodFIter. Figure 3 presents the error of the fitted angular velocity Chebyshev polynomial, as compared with the true angular velocity (24). The comparison is made at one tenth of the original sampling interval, namely, 1000 Hz.

Figure 4 plots the fast RodFIter's attitude error for the truncation order $n_T = n+1$, where the iteration times are set to 7. The attitude error of the non-truncated/original RodFIter in [1] is also given for easy comparison. The error discrepancy between the fast RodFIter and the original RodFIter is hardly discerned, which shows that the Chebyshev polynomial truncation (17) has expectedly little negative effect on the attitude accuracy. Figure 5 compares the absolute value of the non-truncated coefficients in (15) and the truncated coefficients in (17) of the corresponding Rodrigues vector at each iteration, in which the coefficients at the last iteration are highlighted by thicker lines. For clarity, the non-truncated coefficients in the first three iterations are only presented. For the non-truncated version, the order of Chebyshev polynomial is $m_l = (2^3-1)(7+1) = 56$; while the polynomial order is only $n+1=8$ for the truncated RodFIter. It is clear from the left subfigure that the Chebyshev polynomial's coefficients constantly decrease in magnitude with increasing order, which makes it possible to speed up the original RodFIter by truncation. In the right subfigure, the coefficients of the truncated Chebyshev polynomial are well below $10^{-15}$ in magnitude. Figure 6 further compares the coefficients of the Chebyshev polynomial at the 7[th] iteration for three different truncation degrees ($n$-1, $n$+1 and $n$+3). The Chebyshev polynomial coefficients of the true Rodrigues vector are also plotted for reference (See Appendix for calculation details). As Theorem 2 predicts, the coefficients totally agree very well with each other at the same polynomial order. The inconsistency of the true coefficients in the right-bottom corner of Fig. 6 is owed to machine precision. It is apparent that the truncation error of $n_T = n+1$ is compatible with $t\sup|\delta\boldsymbol{\omega}| \approx 10^{-15}$ according to Fig. 3.

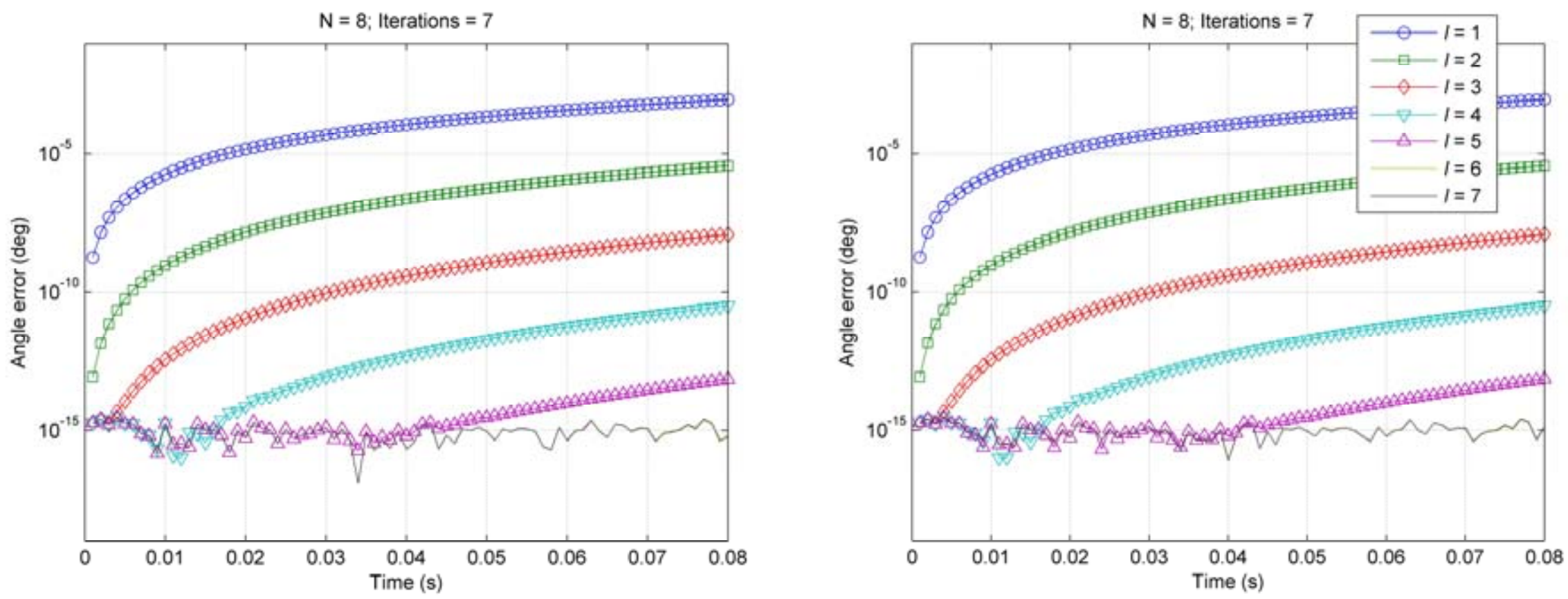

Figure 4. Attitude computation errors by non-truncated/original RodFIter (left, taken from Fig. 3 in [1]) and truncated RodFIter (right).

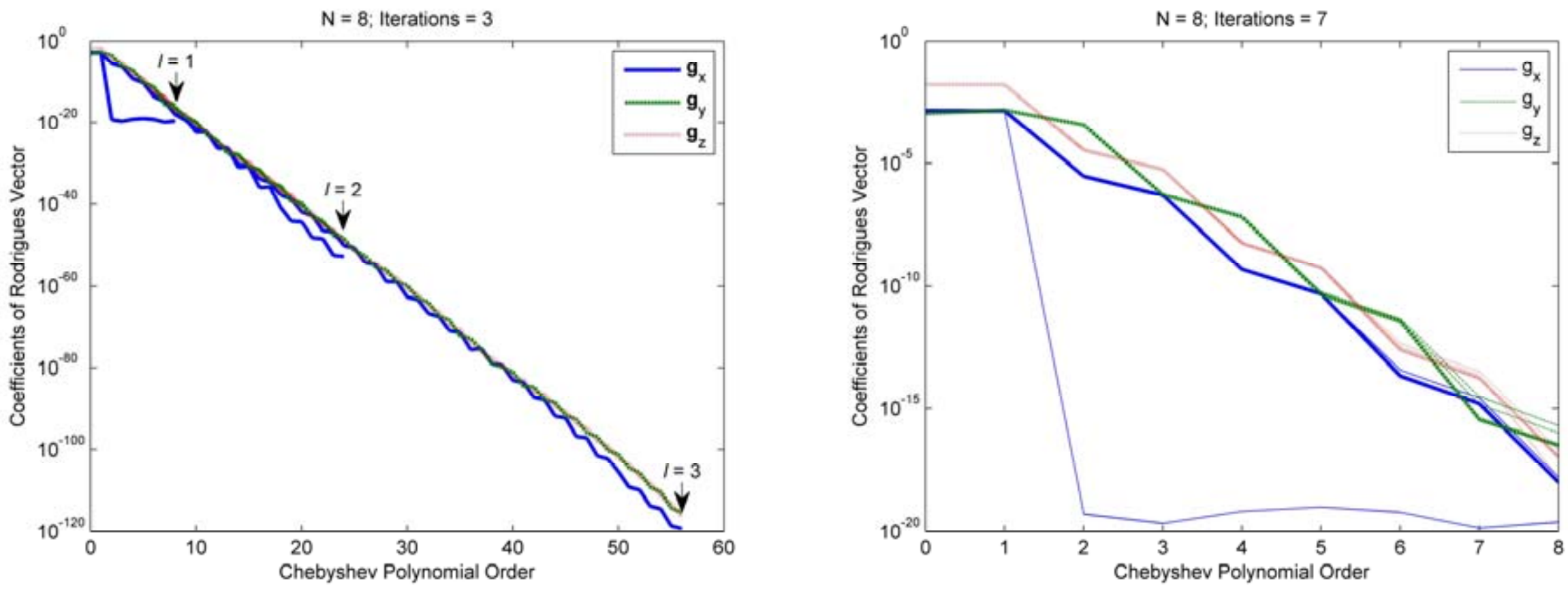

Figure 5. Magnitude of non-truncated coefficients in (15) and truncated coefficients in (17) of the Rodrigues vector at each iteration. Coefficients at last iteration are plotted in thicker lines. Non-truncated coefficients in the first three iterations are only presented for clarity.

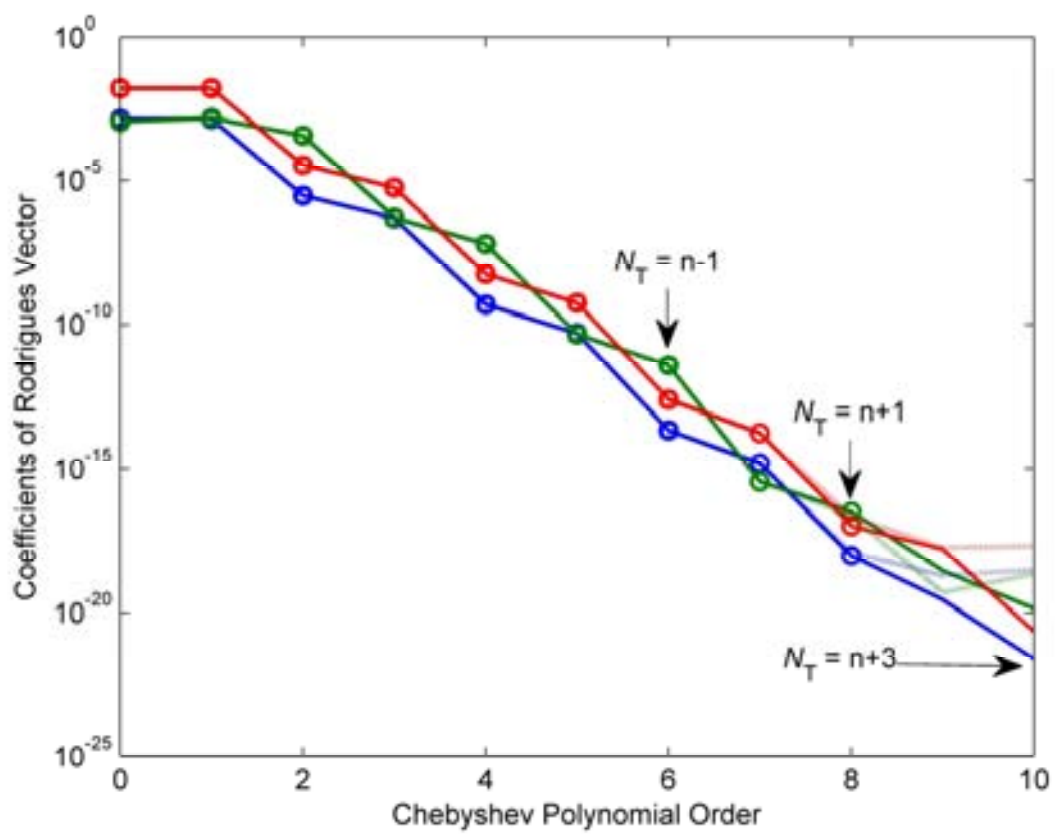

Figure 6. Coefficient comparison of Chebyshev polynomial at the 7[th] iteration for three different truncation degrees, namely, $N_T = n-1$ (solid line decorated by squares), $n+1$ (solid line decorated by circles), and $n+3$ (solid line). The dotted line denotes Chebyshev polynomial coefficients of true Rodrigues vector.

Figure 7 presents the attitude computation error of the truncated RodFIter for two seconds, compared with non-truncated RodFIter and the mainstream two-sample algorithm in the practical inertial navigation system. Table I lists the computation time (averaged across 50 runs) of the non-truncated and truncated RodFIters and the mainstream two-sample algorithm. Simulation results show that the truncated RodFIter found herein can reduce the computational

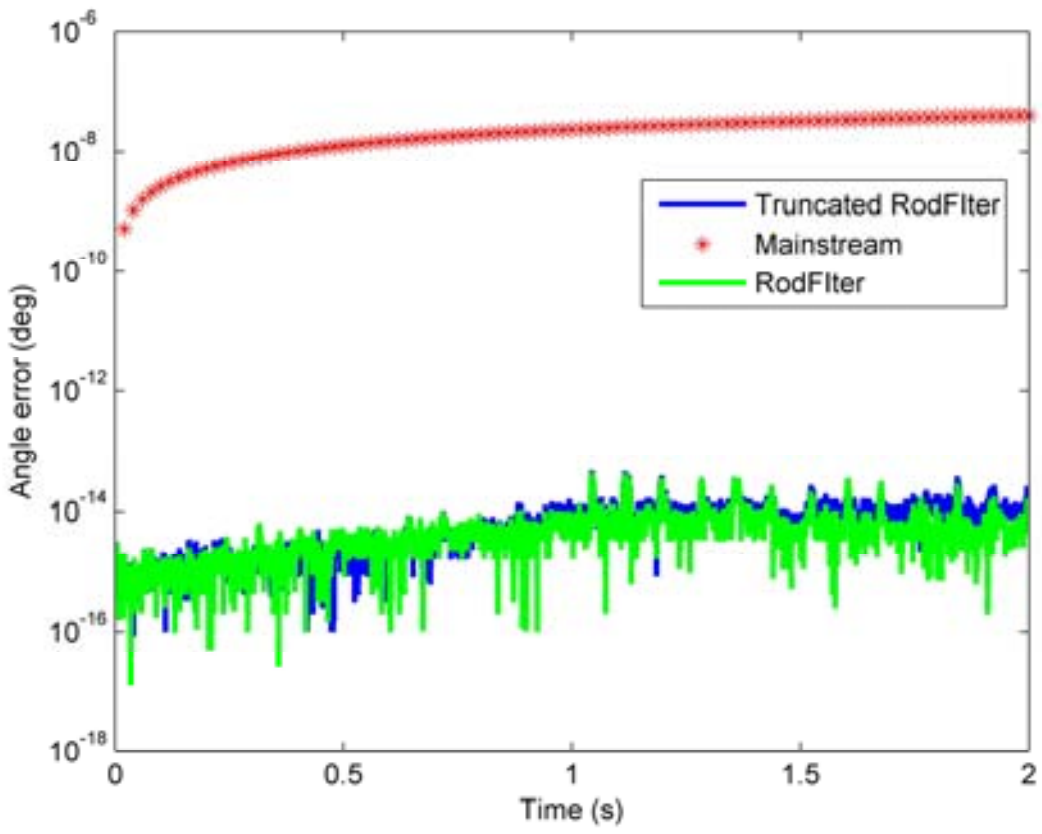

Figure 7. Attitude error comparison of non-truncated and truncated RodFIter ($N$=8) and the mainstream algorithm ($N$=2).

Table I. Comparison of Computation Time (2-s data, sampled 100 Hz)

| **RodFIter** ($N$=8) | **Truncated RodFIter** ($N$=8) | **Mainstream Algorithm** ($N$=2) |
|---|---|---|
| 171s | 0.65s | 0.015s |

Note: Matlab platform on a desktop computer, averaged across 50 runs.

complexity in terms of CPU time by about 260 times over the original RodFIter. Although the truncated RodFIter is still about 40 times slower than the mainstream two-sample algorithm, it is not a problem any more for modern-day computers and an efficient software or hardware implementation might be resorted to further shorten the computational time for those time-sensitive applications.

It should be highlighted that in the comparison of Table I, there is a prejudice in favor of the mainstream algorithm, because it produces attitude results only at the end of incremental update interval while the truncated RodFIter produces attitude results at one tenth of the original sampling interval. In fact, the latter is an analytical reconstruction and thus can produce the attitude in any time scale.

## VI. Discussions and Conclusions

The recapitulation of this paper illustrates the fact that the RodFIter is a promising attitude reconstruction method from discrete gyroscope measurements. This paper reformulates the original RodFIter method in terms of the iterative computation of the Rodrigues vector's Chebyshev polynomial coefficients and then performs the complexity analysis. As the magnitude of the Chebyshev polynomial coefficients decreases along with the polynomial order, a fast version of RodFIter is thus achieved by means of appropriate polynomial truncation at little loss of accuracy. Finally, simulation results show that the computational efficiency is significantly improved by the proposed fast RodFIter. As a result, the simplicity of the computation renders it to be suitable to real-time applications. The idea of economic iterative integration by truncated Chebyshev polynomial could be applied to the above-mentioned high-accurate attitude algorithms and other related engineering problems.

## Appendix

For the Rodrigues vector (25), the incremental Rodrigues vector with respect to the simulation start time is given by [2]

$$\Delta\mathbf{g}(t)=\frac{2\tan\frac{\alpha}{2}}{1+\cos(\Omega t)\tan^2\frac{\alpha}{2}}\left[-\sin(\Omega t)\tan\frac{\alpha}{2}\quad \cos(\Omega t)-1\quad \sin(\Omega t)\right]^T. \tag{28}$$

The incremental Rodrigues vector can be approximated by a Chebyshev polynomial of degree $M$ as $\Delta\mathbf{g}(t)=\sum_{j=0}^{M}\boldsymbol{\beta}_j F_j(\tau)$, where the coefficients are approximately computed as [14]

$$\boldsymbol{\beta}_j\approx\frac{2-\delta_{0j}}{P}\sum_{k=0}^{P-1}\cos\left(\frac{j(k+1/2)\pi}{P}\right)\Delta\mathbf{g}\left(\cos\left(\frac{(k+1/2)\pi}{P}\right)\right), \tag{29}$$

where $\delta_{ij}$ is the Kronecker delta function. Exact coefficients could be obtained if the number of summation terms $P$ approaches infinity.

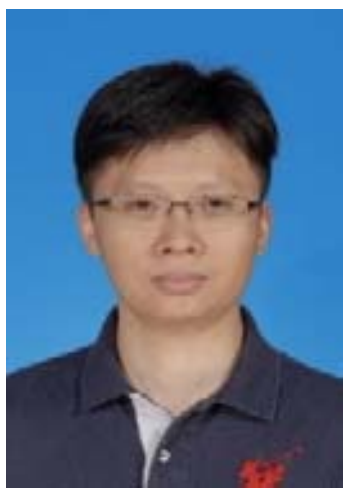

**Yuanxin Wu** (SM'14) received the B.Sc. and Ph.D. degree in navigation from Department of Automatic Control, National University of Defense Technology, in 1998 and 2005, respectively. He was with National University of Defense Technology as a Lecturer (2006-2007) and an Associate Professor (2008-2012), Department of Geomatics Engineering, University of Calgary, Canada, as a visiting Postdoctoral Fellow (2009.2-2010.2) and Central South University as a Professor (2013-2015). He is currently a Professor in School of Electronic Information and Electrical Engineering, Shanghai Jiao Tong University, China. His current research interests include inertial-based navigation system, state estimation, inertial-visual fusion and wearable human motion sensing.

He was the recipient of National Excellent Doctoral Dissertation (2008), New Century Excellent Talents in University (2010), Fok Ying Tung Education Fellowship (2012), NSFC Award for Excellent Young Scientists (2014), Natural Science and Technology Award in University (2008, 2016) and Elsevier's Most Cited Chinese Researchers (Aerospace Engineering, 2015-2017). He serves as an Associate Editor for The *Journal of Navigation* and *IEEE Trans. on Aerospace and Electronic Systems*. He is Senior Member of the IEEE.

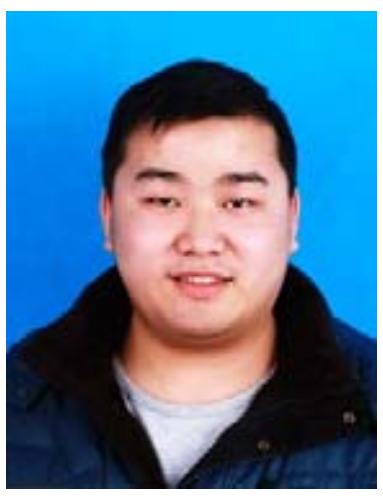

**Qi Cai** received the B.Sc. and M.S. degrees respectively in School of Geosciences and Info-Physics and School of Aeronautics and Astronautics, Central South University, in 2015 and 2018. He is now pursuing his Ph.D. degree in School of Electronic Information and Electrical Engineering, Shanghai Jiao Tong University. His research interests include geodesy, computer vision and inertial-visual fusion for navigation.

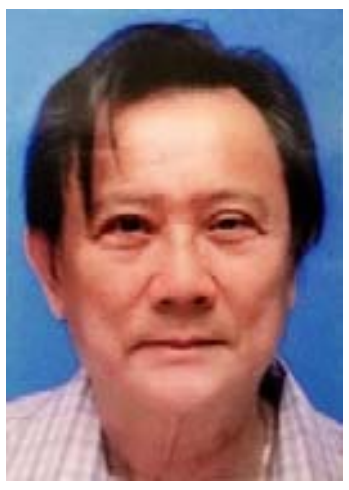

**Trieu-Kien Truong** (M'82-SM'83-F'99-LF'13) received the B.Sc. degree in Electrical Engineering from national Cheng Kung University, Taiwan in 1967, the M.S. degree in Electrical Engineering from Washington University, Missouri in 1971, and the Ph.D. degree in Electrical Engineering from the University of Southern California, Los Angeles in 1976. From 1975 to 1992, he was a Senior Member of Technical Staff (E6) in the Communication System Research Section of the JPL, Pasadema, California, from 1976 to 1995. He was an adjunct professor in the Communications Science Institute, Department of Electrical Engineering-Systems at USC, and was a consultant to Department of Radiology, Memorial Hospital of Long Beach. Also, from 1995 to 2010, he was a Chair Professor and Dean of the college of Electrical and Information Engineering, I-Shou University, Taiwan. Currently, he is a Visiting Distinguished Chair Professor at I-Shou University and is a visiting Chair Professor in School of Electronic Information and Electrical Engineering of Shanghai Jiao Tong University.

Dr. Truong is the recipient of many honors, including 23 NASA awards for outstanding technical contributions. He hold many patents in different fields and is the author of more than 200 journal papers, among them about 26 papers published in IEEE Trans on Information Theory. His main research interests include error correcting code, VLSI architecture design, communication systems, signal and image processing, SAR digital processor, aerial images, and CT-aided robotic stereotaxis system. He is Life Fellow of the IEEE.